\documentclass[10pt,twocolumn,letterpaper]{article}

\usepackage{wacv}
\usepackage{times}
\usepackage{epsfig}
\usepackage{graphicx}
\usepackage{amsmath}
\usepackage{amssymb}

\usepackage{booktabs}
\usepackage{xcolor,colortbl}
\usepackage[super]{nth}
\usepackage{subcaption}
\usepackage[acronym]{glossaries}
\usepackage{array,multirow}

\usepackage{pifont}% http://ctan.org/pkg/pifont
\newcommand{\cmark}{\ding{51}}%
\newcommand{\xmark}{\ding{55}}%

\newacronym{evs}{EVS}{Efficient Video Segmentation}
\newacronym{iam}{IAM}{Inconsistencies Attention Module}
\newacronym{cnn}{CNN}{Convolutional Neural Network}
\newacronym{fcn}{FCN}{Fully Convolutional Network}
\newacronym{crf}{CRF}{Conditional Random Fields}
\newacronym{of}{OF}{Optical Flow Module}
\newacronym{dis}{DIS}{Dense Inverse Search}

\wacvfinalcopy % *** Uncomment this line for the final submission

\def\wacvPaperID{432} % *** Enter the wacv Paper ID here

\ifwacvfinal\fi
\begin{document}

%%%%%%%%% TITLE
\title{Efficient Video Semantic Segmentation with Labels Propagation and Refinement}

% Authors at the same institution
\author{Matthieu Paul \hspace{1cm} Christoph Mayer \hspace{1cm} Luc Van Gool \hspace{1cm} Radu Timofte\\
{\tt\small \{paulma,chmayer,vangool,timofter\}@vision.ee.ethz.ch}\\
Computer Vision Lab, ETH Z{\"u}rich, Switzerland\\
}

% \author{Matthieu Paul \hspace{2cm} Christoph Mayer \hspace{2cm} Radu Timofte \hspace{2cm} Luc Van Gool \\
% {\tt\scriptsize paulma@vision.ee.ethz.ch}\hspace{1cm}
% {\tt\scriptsize chmayer@vision.ee.ethz.ch}\hspace{1cm}
% {\tt\scriptsize timofter@vision.ee.ethz.ch}\hspace{1cm}
% {\tt\scriptsize vangool@vision.ee.ethz.ch}\\
% ETH Z{\"u}rich, Switzerland\\
% }

% \author{Matthieu Paul \hspace{2cm} Christoph Mayer \hspace{2cm} Radu Timofte \hspace{2cm} Luc Van Gool \\
% ETH Z{\"u}rich, Switzerland\\
% {\tt\small paulma@vision.ee.ethz.ch}
% }

% \author{Matthieu Paul\\ 
% ETH Z{\"u}rich, Switzerland\\
% {\tt\scriptsize paulma@vision.ee.ethz.ch}
% % For a paper whose authors are all at the same institution,
% % omit the following lines up until the closing ``}''.
% % Additional authors and addresses can be added with ``\and'',
% % just like the second author.
% % To save space, use either the email address or home page, not both
% \and
% Christoph Mayer\\
% ETH Z{\"u}rich, Switzerland \\
% {\tt\scriptsize chmayer@vision.ee.ethz.ch}
% \and
% Radu Timofte\\
% ETH Z{\"u}rich, Switzerland \\
% {\tt\scriptsize timofter@vision.ee.ethz.ch}
% \and
% Luc Van Gool\\ 
% ETH Z{\"u}rich, Switzerland\\
% {\tt\scriptsize vangool@vision.ee.ethz.ch}
% }

\maketitle
\ifwacvfinal\thispagestyle{empty}\fi

%%%%%%%%% ABSTRACT
\begin{abstract}
This paper tackles the problem of real-time semantic segmentation of high definition videos using a hybrid GPU-CPU approach. We propose an \gls{evs} pipeline that combines: 

%(i) On the GPU, two coexisting deep models: a "heavy" deep neural network that is used to predict dense semantic labels from scratch and a lighter network designed to improve predictions from the previous frames. The latter refines the semantic predictions in regions that were identified as difficult to propagate, thanks to a fast \acrlong{iam}.

(i) On the CPU, a very fast optical flow method, that is used to exploit the temporal aspect of the video and propagate semantic information from one frame to the next. It runs in parallel with the GPU.

(ii) On the GPU, two Convolutional Neural Networks: A main segmentation network that is used to predict dense semantic labels from scratch, and a Refiner that is designed to improve predictions from previous frames with the help of a fast \gls{iam}. The latter can identify regions that cannot be propagated accurately.

%in combination with a fast \gls{iam}, is proposed to refine the semantic predictions propagated from frames to frames, which can be distorted over time. 

We suggest several operating points depending on the desired frame rate and accuracy. Our pipeline achieves accuracy levels competitive to the existing real-time methods for semantic image segmentation (mIoU above 60\%), while achieving much higher frame rates. On the popular Cityscapes dataset with high resolution frames ($2048\times1024$), the proposed operating points range from 80 to 1000 Hz on a single GPU and CPU.

{\bf Keywords:} Real-Time, Video Semantic Segmentation, Optical flow, Propagation, Refinement

\end{abstract}

%%%%%%%%% BODY TEXT
\section{Introduction}

\begin{figure}[t]
\begin{center}
\includegraphics[width=\linewidth]{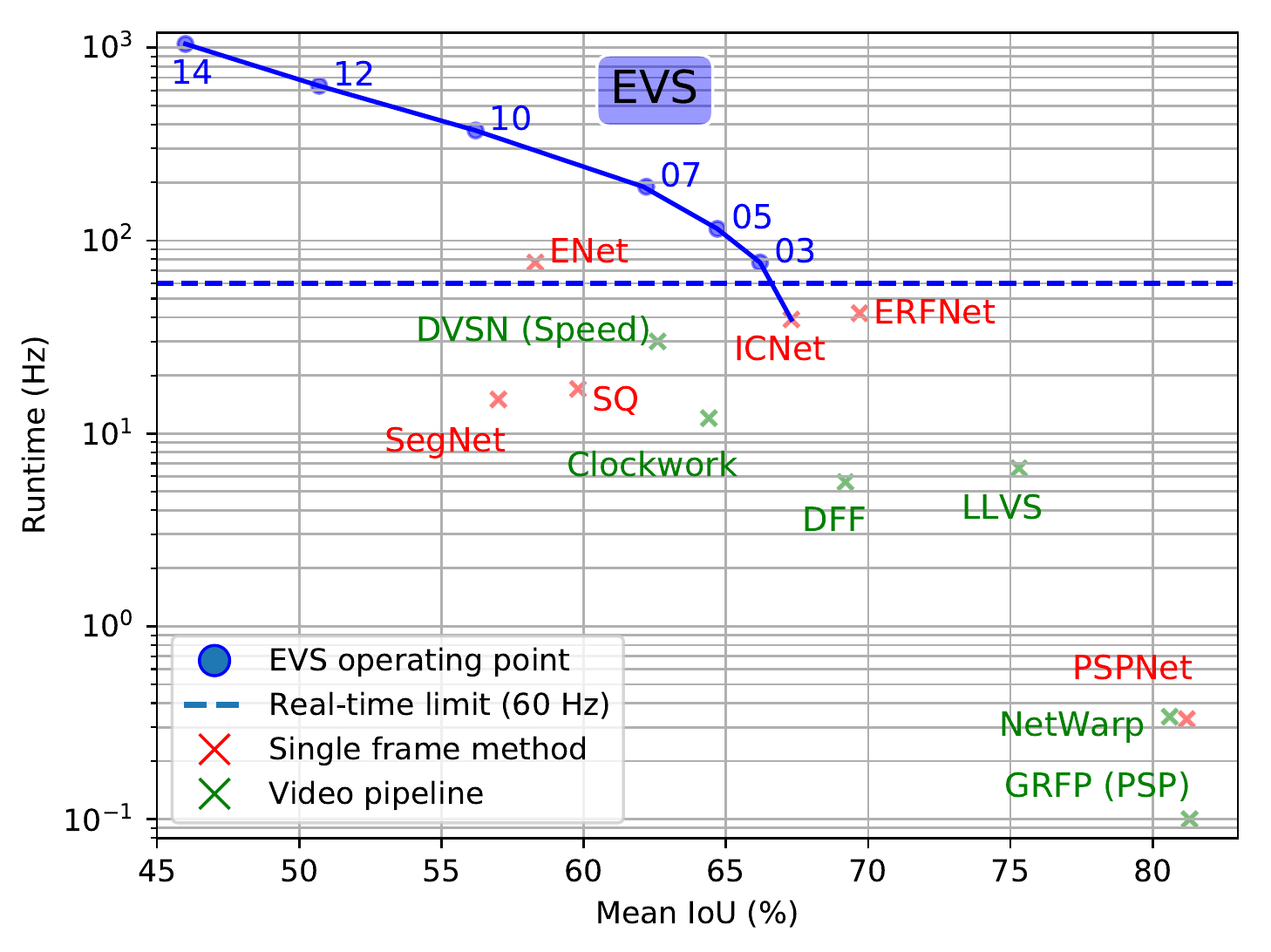}
\end{center}
\caption{Comparison between our EVS pipeline and state of the art methods on the Cityscapes~\cite{Cordts2016Cityscapes} dataset with input resolution $2048\times1024$. Table \ref{tab:op_points} provides more operating points and comparisons, from 1 Hz to 1000 Hz.}
\label{fig:op_points}
\end{figure}

A lot of efforts have been made in semantic segmentation over the past years. Yet, while segmentation accuracy reached astonishing levels, little focus has been put on making it usable in real-time scenarios. Achieving very fast semantic segmentation would have many advantages, especially when used as an additional building block for other computer vision tasks related to real-time scene understanding. Particularly in the context of real-world scenarios for industrial or commercial cases such as augmented reality, autonomous driving, autonomous flying, etc.

Video scene understanding is already a wide and active research topic, especially in accurate object instances tracking and segmentation. However, in the context of real-time video semantic segmentation, fewer efforts have been put into exploiting the temporal information as a mean to decrease inference time. When used, this temporal aspect is in most methods used as an additional information to improve either the accuracy of the predictions or their consistency over time, at the cost of additional runtime.

On the contrary, the focus of this paper is to use temporal information as a way to minimize the inference time for each frame as much as possible, while limiting the drop in accuracy resulting from the reduced computations. The baseline we use runs at around 40 Hz on a frame resolution of $2048\times1024$. Our \gls{evs} pipeline defines several operating points among which the speedup factor varies from $\times2$ to $\times27$ on the same resolution.

The proposed pipeline uses ICNet~\cite{Zhao_2018_ECCV} as the main prediction network, since it is the current state of the art in terms of trade-off between accuracy and performance for single frame processing. To compute the dense optical flow, we use \gls{dis}~\cite{kroegerECCV2016} as it is the current state of the art in terms of computational efficiency on the CPU. Dense optical flow plays a key role in our pipeline, as it can run on the CPU in parallel with the GPU at a much higher frame rate than the prediction network. This information is then used to:

\begin{itemize}  
 \item[-] Warp the semantic information from one frame to the next, both of high level predictions and low level contextual features. This warped semantics is used as input for the Refiner that will improve the labels prediction for the current frame. 
 \item[-] Feed the \gls{iam} to focus the refinement on regions where the optical flow is unreliable (typically thin and/or moving objects boundaries), by computing the forward-backward consistency of the propagated labels.
\end{itemize}

%-------------------------------------------------------------------------

\subsection{Contributions}

Since semantic segmentation is crucial for video scene understanding, we aim at pushing the limits of this field through the following contributions, with a focus on efficiency and frame rate.

First, our hybrid \gls{evs} pipeline balances the workload between GPU and CPU. They work in parallel, either computing semantic predictions or propagating them from frame to frame using optical flow, instead of having one large pipeline running fully on the GPU. Running the optical flow directly on the CPU decreases the workload on the GPU and leads to a massive reduction in computation time. Our goal is to establish new standards in terms of speed for real-time video semantic segmentation while preserving a sound segmentation quality.

% The direct advantage of offloading the optical flow computation on the CPU is a massive gain in computation time. This allows us to push the boundaries of video semantic segmentation in terms of speed by redefining the notion of "real-time" for that task, showing that extremely high frame rates can be attained on a single GPU and CPU pair, while still achieving reasonable accuracy.

Furthermore, we introduce a fast \gls{iam} and a Refiner that work together to refine the propagated predictions of the main segmentation network to better match the current frame. Our versatile design allows running our pipeline in various operating modes, trading-off speed versus segmentation quality.

% Secondly, we show that having a dense optical flow available from the CPU can be exploited to do less work on the GPU when needed. We introduce a fast \gls{iam} and a "light" \gls{cnn} that can work together to refine the predictions generated by the "heavy" network and propagated over time. We suggest different operating modes for our pipeline that can be used to put more emphasis on speed or accuracy, depending on the desired output.

%-------------------------------------------------------------------------

\section{Related Work}

The most straightforward way to perform video semantic segmentation is to simply run image semantic segmentation on each frame. Although this approach is rather slow, it leads to a natural baseline to assess the quality of video segmentation methods. Furthermore, we review recent trends and ideas in video segmentation. As our proposed method combines semantic image segmentation with optical flow, we review different methods extracting optical flow between consecutive frames using traditional or deep learning-based methods.

\subsection{Image Semantic Segmentation}

Semantic image segmentation aims at assigning a class label to each pixel of a given image. The recent advances in deep learning~\cite{Krizhevsky:2012:ICD:2999134.2999257,DBLP:journals/corr/WuSH16e} lead to fast progress in semantic image segmentation~\cite{DBLP:conf/cvpr/LongSD15, Liu2015SemanticIS, DBLP:journals/pami/ChenPKMY18}. Most of the  state-of-the-art methods~\cite{Chen_2018_ECCV,Zhao_2017_CVPR} are based on \glspl{fcn}~\cite{DBLP:conf/cvpr/LongSD15}.
Among these methods are: DeepLabV3+~\cite{Chen_2018_ECCV}, PSPNet~\cite{Zhao_2017_CVPR} or more recently Panoptic FPN~\cite{DBLP:journals/corr/abs-1901-02446}. These methods concentrate mainly on high quality segmentation masks that require a large amount of parameters and are computationally intensive, \ie inference time of around one second for a high resolution frame ($2048\times1024$).

% Most of these models are based on \glspl{fcn} which makes them computationally intensive as their number of parameters is high. For instance, current state-of-the-art networks for single image semantic segmentation such as DeepLabV3+~\cite{Chen_2018_ECCV}, PSPNet~\cite{Zhao_2017_CVPR} or Panoptic FPN~\cite{DBLP:journals/corr/abs-1901-02446} achieve very high accuracy at the price of high memory consumption and slow processing time: in the order of 1 second to infer labels on a high resolution frame ($2048\times1024$). 

Other methods that focus on reducing computing time and memory footprint obtain more and more attention: SegNet~\cite{Badrinarayanan2016SegNetAD}, SQ~\cite{Treml2016SpeedingUS}, ENet~\cite{DBLP:journals/corr/PaszkeCKC16} and ESPNet~\cite{Mehta_2018_ECCV}.

Combining the best of both worlds, some methods aim at finding good trade-offs between frame rate and accuracy, either from their model (ERFNet~\cite{Romera2018ERFNetER} and ICNet~\cite{Zhao_2018_ECCV}) or by treating differently \emph{complex} and \emph{simple} parts of the image (LC~\cite{Li_2017_CVPR}). These methods achieve faster inference times while preserving a decent segmentation quality. 

% A few other networks started to focus more on memory footprint and computing times, for instance SegNet~\cite{Badrinarayanan2016SegNetAD}, SQ~\cite{Treml2016SpeedingUS}, ENet~\cite{DBLP:journals/corr/PaszkeCKC16} and ESPNet~\cite{Mehta_2018_ECCV}. These methods usually achieve better performances but at the price of drop in accuracy. 

% Recent networks such as ICNet~\cite{Zhao_2018_ECCV} and BiSeNet~\cite{Yu_2018_ECCV} try to find a good trade-off between memory footprint, run-time and precision. They manage to significantly computing costs without degrading too much the segmentation accuracy.  

\subsection{Video Semantic Segmentation}

Compared to semantic image segmentation, developing dedicated video segmentation pipelines is a less explored research track. Applying image segmentation algorithms that operate on each video frame individually is possible. However, specialized methods for videos can exploit temporal information between consecutive frames to enable more reliable predictions or increase the frame rate. 

Early methods tackling video segmentation were extending classical single image segmentation methods with temporally-aware  components: normalized cuts~\cite{Shi:2000:NCI:351581.351611}, tracking~\cite{Lezama2011} or motion segmentation~\cite{Ochs:2014:SMO:2693343.2693376}. 
Recent methods leverage dense optical flow in a more direct way by combining it with Gated Recurrent Units (GRUs) to refine the predictions and add temporal consistency~\cite{tokmakov:hal-01511145, Nilsson_2018_CVPR}. 

% Early methods in that field were extending existing methods with a temporal aspect: normalized cuts~\cite{Shi:2000:NCI:351581.351611}, tracking~\cite{Lezama2011}, motion segmentation\cite{Ochs:2014:SMO:2693343.2693376}. Newer methods also use dense optical flow to align consecutive frames potentials for dense \gls{crf}~\cite{Kundu_2016_CVPR} or in combination with Gated Recurrent Units to get a refined and temporally more consistent prediction~\cite{tokmakov:hal-01511145, Nilsson_2018_CVPR}. 

\begin{figure*}[t]
\begin{center}
\includegraphics[width=\linewidth]{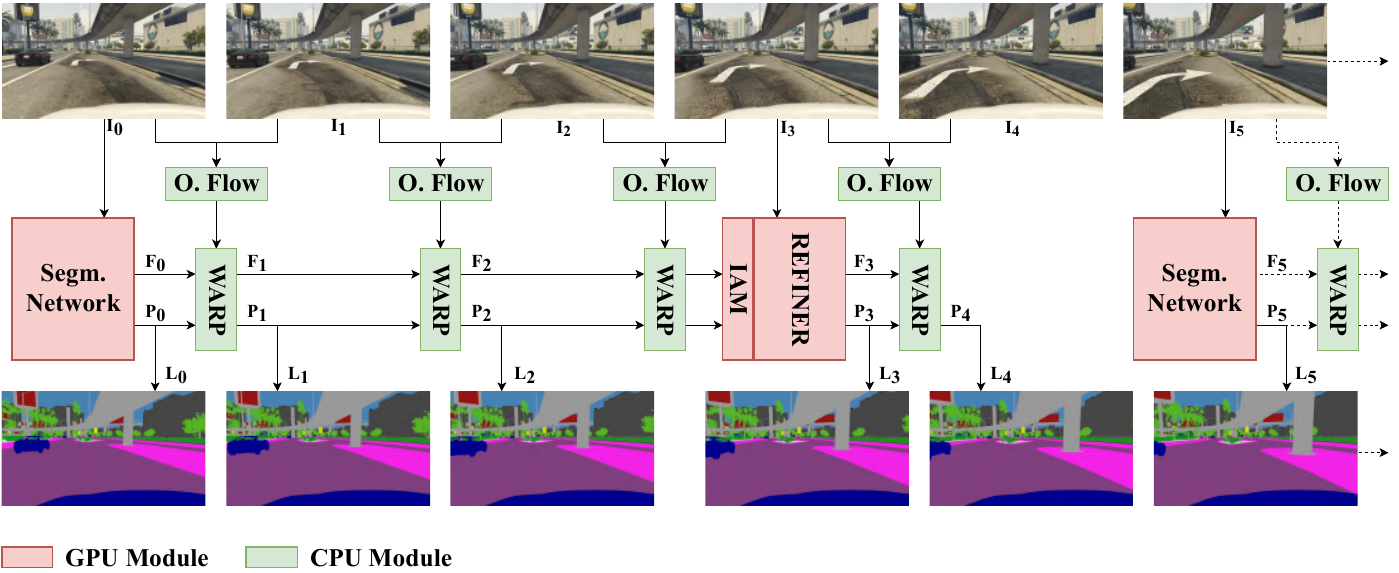}
\end{center}
\caption{Full pipeline overview with an input video stream $(\boldsymbol{I_0, I_1, ...})$ and the corresponding output labels $(\boldsymbol{L_0, L_1, ...})$. The predicted probabilities $\boldsymbol{P_i}$, labels $\boldsymbol{L_i}$ and deep features $\boldsymbol{F_i}$ are propagated with the corresponding dense optical flow.}
\label{fig:network_overview}
\end{figure*}

In particular, some methods aim at reducing inference times by embedding the temporal aspect in their structure by using LSTM~\cite{Mahasseni2017BudgetAwareDS}, or by selecting key frames to fully segment. Clockwork~\cite{10.1007/978-3-319-49409-8_69} authors observe that intermediate representations within a network change only slowly in most videos. Therefore, they propose to schedule features computation for key frames only and share features in between. LLVS~\cite{Li_2018_CVPR} and DVSN~\cite{Xu_2018_CVPR} try to further optimize scheduling depending on frame content.

% Clockwork~\cite{10.1007/978-3-319-49409-8_69} for video semantic segmentation on the other hand is exploiting the intermediate representations in the network, making the observation that those features change slowly in a video. This allows them to recompute features only when needed.

Another family of methods uses the geometrical structure of the 3D scene to improve the segmentation quality. There, 3D point clouds obtained from visual odometry or stereo-vision approaches add additional information that allows more reliable predictions~\cite{Brostow2008SegmentationAR, floros12, Sengupta2013Urban3S, 10.1007/978-3-319-10599-4_45}.

% Another range of methods rely on the geometrical structure of the scene, where semantic video segmentation accuracy is improved by the additional computation of 3D point clouds obtained by visual odometry or stereo-vision algorithms~\cite{Brostow2008SegmentationAR, floros12, Sengupta2013Urban3S, 10.1007/978-3-319-10599-4_45}.
% Semantic segmentation pipelines tailored for video processing are much scarcer in comparison to semantic segmentation for single images. Instead of computing a full semantic segmentation for each frame, some methods try to leverage the temporal information contained in consecutive frames, to perform either video object segmentation or video semantic segmentation. One of the main issues in that research field is the lack of densely annotated data: Dataset usually provide some video snippets, but they are rarely densely annotated~\cite{Brostow2009SemanticOC, Geiger2013IJRR, Cordts2016Cityscapes}, or tailored towards video object segmentation~\cite{Perazzi2016}.

One of the major challenges in video segmentation remains the massive amount of data that deep \glspl{cnn} require for training. Already, producing annotations for semantic image segmentation is costly. In this case, ensuring diversity in a video segmentation data set demands many different video sequences each consisting of hundreds of frames even for very short movies, leading to thousands of frames that should be annotated. Thus, existing data sets for video segmentation are either sparsely annotated~\cite{Brostow2009SemanticOC, Geiger2013IJRR, Cordts2016Cityscapes}, \ie not every frame is labelled, or the segmentation task is simplified such that annotation is cheaper, \ie single object segmentation~\cite{Perazzi2016}. In our case, we avoid this pitfall by relying on a network which is trained on single images. Only synthetic data sets such as GTA5~\cite{Richter_2016_ECCV} or Sintel~\cite{Butler:ECCV:2012} overcome that annotation limitation.

\subsection{Optical Flow}

Traditional optical flow methods such as Lucas-Kanade~\cite{DBLP:conf/ijcai/LucasK85} or Gunnar-Farneback~\cite{DBLP:conf/scia/Farneback03}, recently started to compete with new deep learning approaches: FlowNet~\cite{DFIB15, IMSKDB17}, MPNet~\cite{Tokmakov17} and SegFlow~\cite{Cheng_ICCV_2017} produce very accurate flow estimates, but are rather expensive and slow and run on the GPU. As a result, deep video semantic segmentation pipelines using optical flow usually improve marginally their accuracy or temporal consistency, while increasing substantially their inference time: NetWarp~\cite{gadde2017semantic}, GRFP~\cite{Nilsson_2018_CVPR} or DFF~\cite{zhu17dff}. 
In contrast, when aiming at fast and efficient video segmentation, \gls{dis}~\cite{kroegerECCV2016} is among the most suitable candidates. \gls{dis} achieves much higher frame rates than deep optical flow methods and runs on CPU, which gives more flexibility to choose between speed and accuracy by selecting different operating points.

\section{Efficient Video Segmentation Pipeline}

\subsection{Full Pipeline Overview}

Our pipeline consists of five main components that process the video stream jointly, see Figure ~\ref{fig:network_overview}. The GPU holds a segmentation network and a Refiner with \gls{iam}, while the CPU is responsible for computing in parallel the optical flow and for warping the \gls{cnn} features and predictions.

The dense optical flow is computed for each pair of the  consecutive frames. It enables the forward and backward remapping of semantic information extracted by the deep networks. The \gls{iam} is responsible for computing the inconsistencies that remapping reveals. It provides this information to the Refiner, which then corrects mistakes caused by warping around inconsistent areas, i.e. where the optical flow is not reliable.

In the best case, the flow will be consistent and the prediction of the next frame will simply be the previous prediction warped by using optical flow. In most cases, the lack of flow consistency in some regions of the image (sudden changes in brightness, occlusions, multiple fast motions, etc.) will be recovered by the Refiner, while the other prediction of other regions will still be derived from the previous prediction to increase temporal consistency.

\subsection{Semantic Segmentation Network}

The segmentation network in our pipeline is responsible for providing a full frame semantic segmentation. We want to emphasize that any deep framework can be used within our framework, leaving space for improvements when better networks are developed. For this work, we choose to use ICNet~\cite{Zhao_2018_ECCV} because of its excellent trade-off between accuracy and speed: 67\% mIoU at $\sim40$ Hz on the popular Cityscapes~\cite{Cordts2016Cityscapes} dataset.

\subsection{Optical Flow and Semantics Propagation}

The advent of deep learning brought many optical flow methods to impressive quality levels while focusing less on the computational efficiency. However, we require a fast but still accurate dense optical flow. \gls{dis} Flow~\cite{kroegerECCV2016} matches perfectly this requirement and has the advantage of producing a reasonable dense flow at a very high frame rate while running on the CPU. Thus, it allows to save the GPU resources for other tasks.

Dense optical flow provides for each pixel $(x,y)$ of the image a flow in each dimension $F_{xy}^{1\rightarrow 2}(x,y)$, between two consecutive frames $I_1$ and $I_2$. The mapping between $I_1$ and $I_2$ can be written for each dimension as follows:

\begin{align}
\begin{split}
\label{eq:mapping_fct}
M_x(x,y) = I_2(x,y) - F_x^{1\rightarrow2}(x,y)
\\
M_y(x,y) = I_2(x,y) - F_y^{1\rightarrow2}(x,y)
\end{split}
\end{align}

Using Eq.~\eqref{eq:mapping_fct} then allows to produce image $I_2$ solely by remapping the pixels from image $I_1$. For non-integer valued coordinates, using the nearest neighbours interpolation results in a valid remapped image: 

\begin{equation} 
I_2(x, y) = I_1(M_x(x,y), M_y(x,y))
\end{equation}

We want to emphasize that such a mapping is fast to perform because it is highly parallelizable on CPU. In our pipeline, it is used to quickly remap both the predictions and the low level \gls{cnn} features from one frame to the next. These features represent the slow changing contextual information of the scene. The predictions can also be remapped backward such that the \gls{iam} is able to compute the inconsistencies (Figure~\ref{fig:iam_module}).

\subsection{Inconsistencies Attention Module}

The \gls{iam} is working together with the Refiner. It is designed such that it is lightweight and able to focus the attention of the refinement on regions where the optical flow is inconsistent. The inputs are:
\begin{itemize}  
 \item[-] $L_\textrm{F}$: the labels predicted for the current frame, obtained by warping the previous frame labels forward.
 \item[-] $L_\textrm{BF}$: the labels predicted for the current frame, obtained by warping the labels backward and then forward $L_F$.
 \item[-] $P_\textrm{refiner}$: the predicted probabilities for each class by the Refiner.
 \item[-] $P_\textrm{warped}$: the predicted probabilities warped using the optical flow.
\end{itemize}

\begin{figure}[t]
\begin{center}
\includegraphics[width=0.8\linewidth]{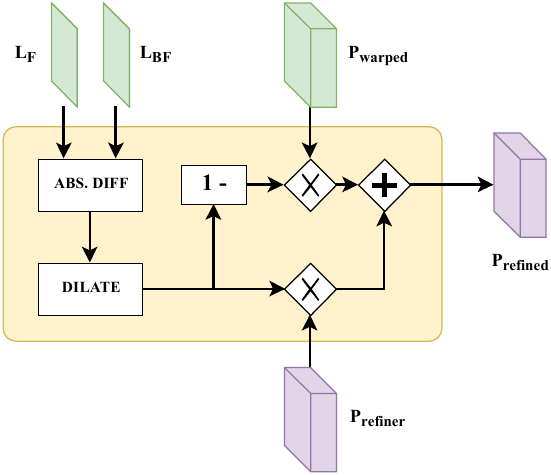}
\end{center}
   \caption{Inconsistencies Attention Module.}
\label{fig:iam_module}
\end{figure}

As a first step, the module computes a probability map $M_i$ that represents the forward-backward inconsistencies of the optical flow for the given input frames. For every pixel $(m,n)$ where $L_\textrm{F}$ and $L_\textrm{BF}$ are different, the probability is considered to be maximal because the flow is unreliable. All other pixels are considered to be reliable:

\begin{equation}
M_i(m,n) = \begin{cases}
1.0 &\text{if $L_\textrm{F}(m,n) \neq L_\textrm{BF}(m,n)$}\\
0.0 &\text{otherwise}
\end{cases}
\end{equation}

As a second step, this binary mask is dilated and smoothed to engulf the surrounding areas of the inconsistencies and to let the Refiner act on them, as the predictions in these regions are more likely to be wrongly propagated by the optical flow.

Finally, the predicted probabilities for each pixel are weighted differently between the warped prediction and the prediction of the Refiner. If $P_\textrm{refiner}$ is the prediction of the Refiner and $P_\textrm{warped}$ is the previous prediction warped to the current frame using the optical flow, the final refined prediction $P_\textrm{refined}$ is defined as the sum of the Hadamard products:

\begin{equation}
P_\textrm{refined} = M_i \circ P_\textrm{refiner} + (1 - M_i) \circ P_\textrm{warped}
\end{equation}

As shown in Figure~\ref{fig:iam_module}, the module only consists of lightweight operations for a GPU, especially since the inputs and outputs are processed at a resolution of $512\times256$.

\subsection{Refiner}

A carefully performed benchmark of the branches in the ICNet architecture shows that even though the network is designed to limit the heavy computations on the lowest resolution to limit the inference time per frame, almost half of that time is spent only on building low level features (see Figure~\ref{fig:network_bench}). The Refiner is built on the idea that these low level features do not need to be recomputed every frame in the context of a continuous video stream: due to their resolution, they are changing at the slowest rate over time.

\begin{figure}[t]
\begin{center}
\includegraphics[width=\linewidth]{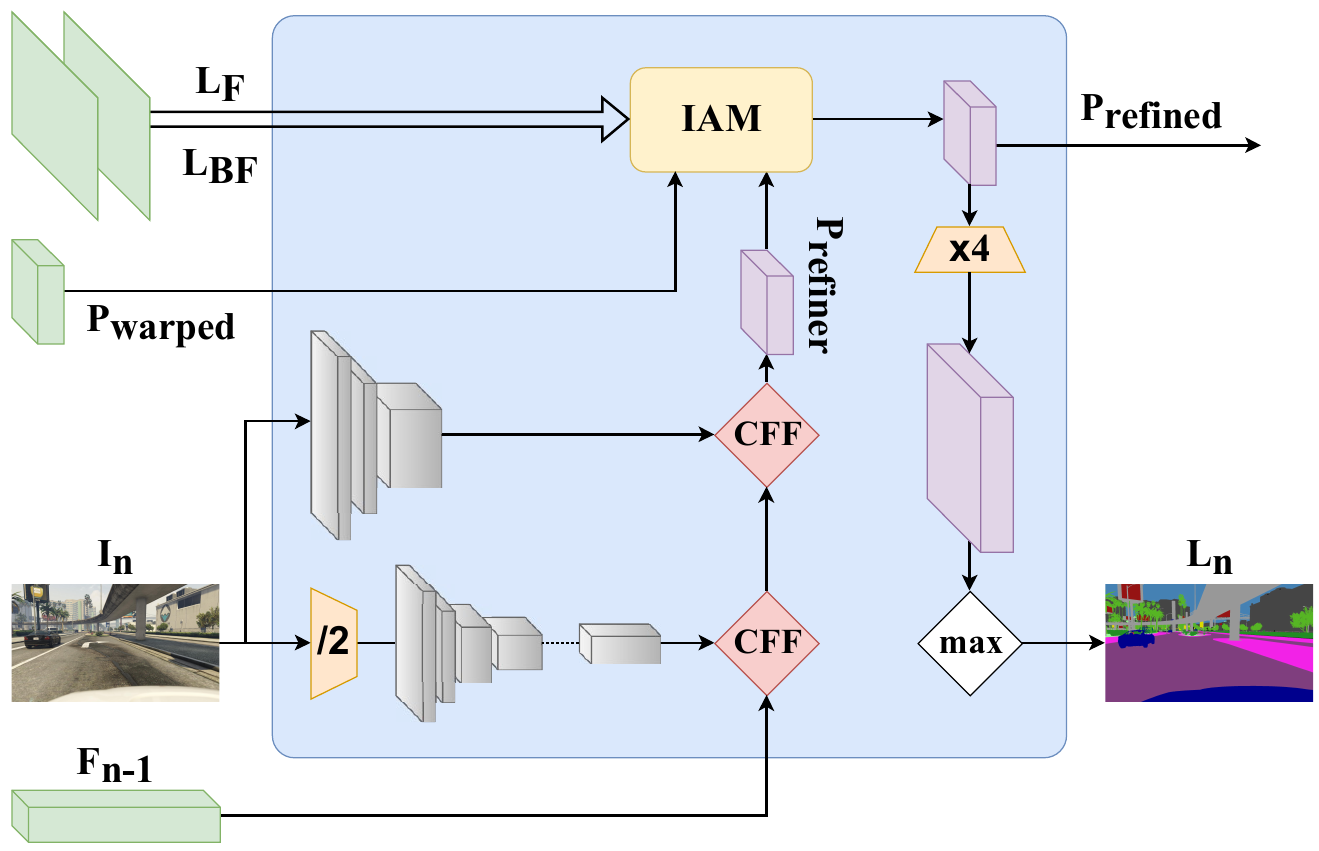}
\end{center}
   \caption{Refiner architecture. For a given input image $\boldsymbol{I_n}$ and warped semantics from the previous frames (green), it generates refined probabilities $\boldsymbol{P_\textrm{refined}}$ and labels $\boldsymbol{L_n}$. CFF stands for "Cascade Feature Fusion", as in ICNet~\cite{Zhao_2018_ECCV}.}
\label{fig:refiner_struct}
\end{figure}

Its task is different from the segmentation network: given pre-aligned low level features from past frames, the Refiner should only compute the higher level features for the new frame, making it shallower. This allows sparing half of the computations that would then otherwise be carried out to extract low level features.

Besides, with the help of the IAM, this refinement is focused only on some areas of the image (see Figure~\ref{fig:refiner_struct}). Using the dense optical flow is reliable for large portions of the image but it causes errors next to object boundaries especially when these objects are thin, moving or new. Thus, using the IAM leads to a better temporal consistency overall, as most of the predicted labels were propagated from one frame to the next.

\section{Experimental evaluation}

\subsection{Setup and Benchmarking Method}

All the benchmarks are done using a single Nvidia Titan Xp GPU, and a Intel Core i7-5930K CPU @ 3.50GHz. The implementation is different from the original ICNet implementation which is written in Caffe and uses a proprietary version of ResNet50. Instead, we use an equivalent implementation in Tensorflow 1.8 and CUDNN 7.1 as the baseline for this paper. The Tensorflow implementation yields almost the same performance and accuracy (67.3\% vs. 67.7\%). All the benchmarks and comparisons in this paper use this Tensorflow implementation. It is worth noting that this is not problematic because the segmentation network of our pipeline can be replaced by any other implementation.

All the following benchmarks and results are produced on Cityscapes~\cite{Cordts2016Cityscapes}, which contains short video snippets of 30 frames at a high resolution ($2048\times1024$) among which the \nth{20} frame contains a fully annotated ground truth mask. All the experiments and results presented are evaluated on the \nth{20} frame with different starting points before it depending on the operating points.

For us, it is important to measure the computation times on the GPU as accurately as possible. Thus, we build a specific probe class based on the publicly available Tensorflow Profiler, which provides the detailed timestamps for each operation on the GPU in JSON format. This data allows us to establish very accurate timings for each part of the network.

Each measurement contains 300 samples from the extracted profiler data. In order to avoid border effects, we measure each sample in middle of the execution of the network. The measurements show that the timings are more varied at the startup time and the initialization of the models.
Nonetheless, following the aforementioned strategy leads to reliable and accurate GPU computation times: the average and median measurements are matching with a small standard deviation, see Figure~\ref{fig:network_bench}.

\subsection{Runtime of the different components}

On the CPU side, warping pixels from one frame to the next using optical flow can be easily parallelized on the CPU. Once split in a $3\times3$ or $4\times4$ grid, all the pixels from a full frame are remapped within a marginal time period ($\sim0.15$ ms on a Intel Core i7-5930K CPU @ 3.50GHz).

\begin{figure}[t]
\begin{center}
\includegraphics[width=\linewidth]{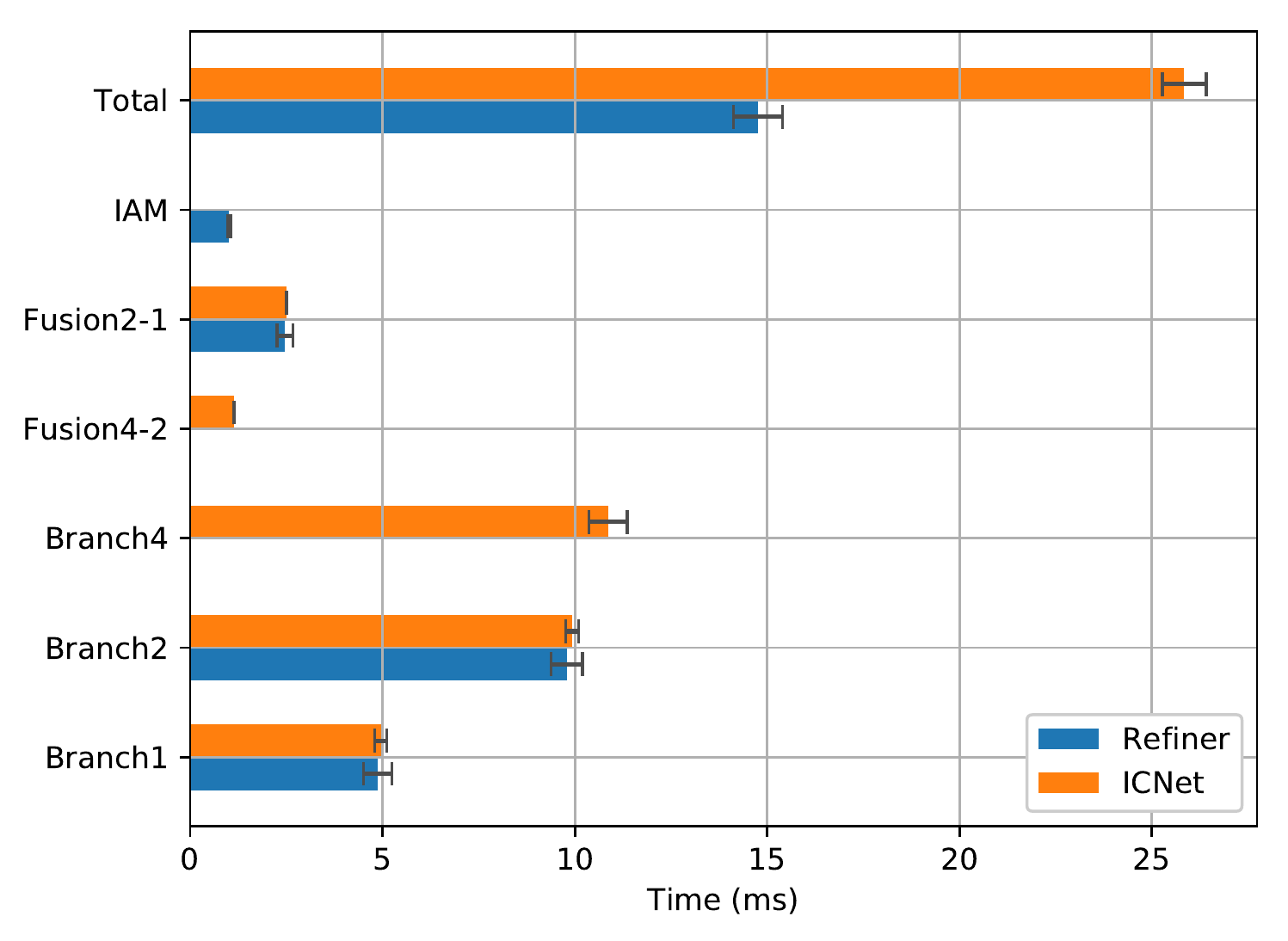}
\end{center}
   \caption{Runtime of the Refiner with \acrshort{iam} compared to ICNet on a single Nvidia Titan Xp.}
\label{fig:network_bench}
\end{figure}

On the GPU side, we have two models: one for the full \gls{cnn} and the other for the Refiner. ICNet~\cite{Zhao_2018_ECCV} is structured around 3 branches: \emph{Branch1} with very few convolutions operating at full resolution, \emph{Branch2} that computes deep features starting from half the resolution, and \emph{Branch4} that goes much deeper at even lower resolution. Figure~\ref{fig:network_bench} shows a speedup of almost two times for the inference time of the Refiner.

\subsection{Optical Flow and Labels Propagation}

\subsubsection{Optical Flow Benchmark}

Several operating points are suggested for \gls{dis}~\cite{kroegerECCV2016}, with a set of parameters that trade off accuracy and runtime. For our experiments, we choose a set of parameters to achieve a small runtime: no variational refinement, finest scale $\theta_f = 2$, patch size $\theta_{ps} = 8$, gradient descent iterations $\theta_{it} = 12$. This allows us to run the optical flow computation on one of the cores of the CPU, on the full frame resolution $2048\times1024$ in less than 5 ms. The goal is to compute the optical flow for five frames on one core, while the segmentation network is working ($\sim25$ ms, see Figure~\ref{fig:network_bench}).

\subsubsection{Influence of the Optical Flow Algorithm}

\begin{figure}[t]
\begin{center}
\includegraphics[width=\linewidth]{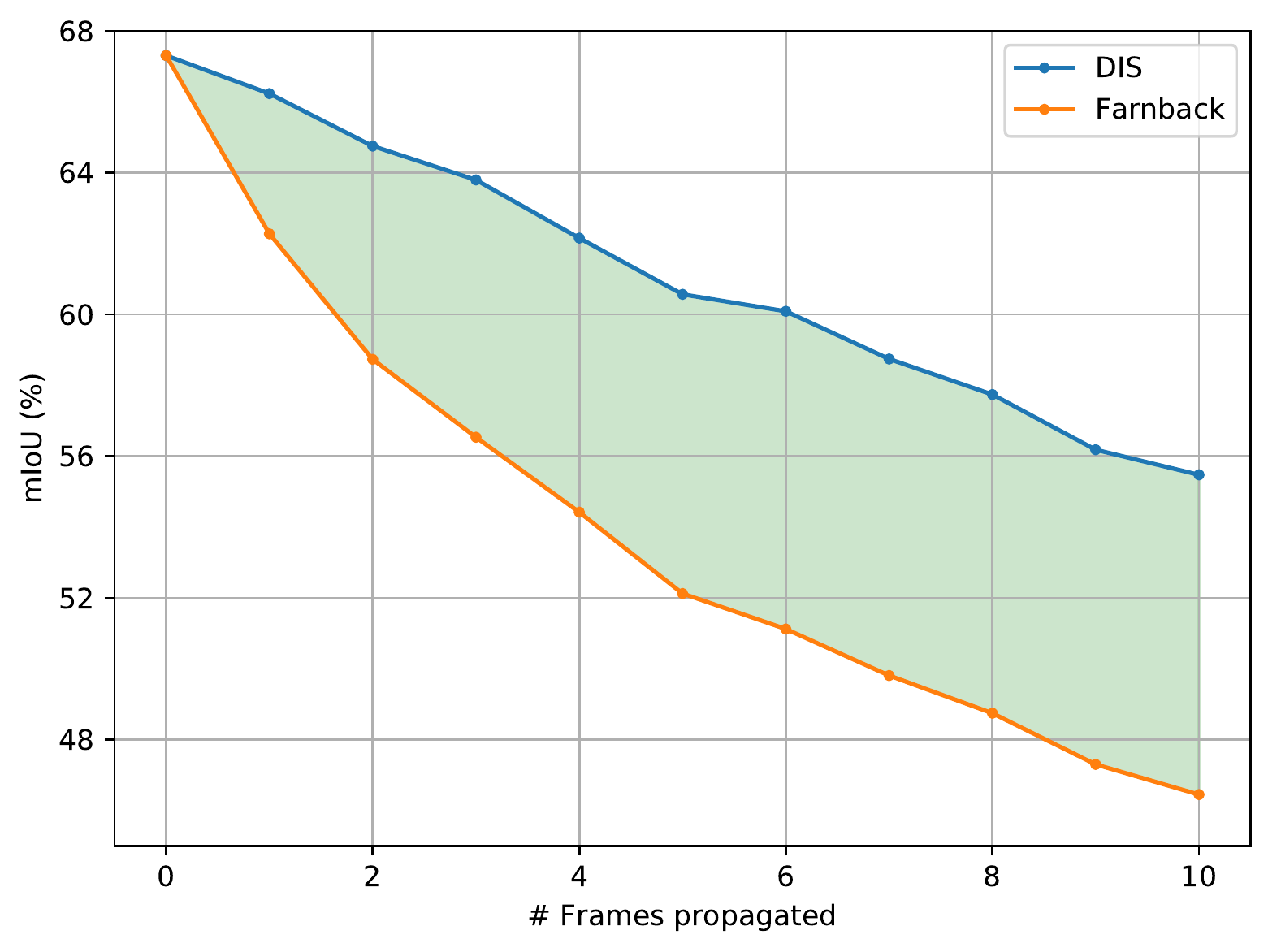}
\end{center}
   \caption{Influence of a pure label propagation on the mIoU for $2048\times1024$, comparison Gunnar-Farneb\"ack vs. \gls{dis}.}
\label{fig:optflow_comp}
\end{figure}

The dense optical flow computation is of paramount importance to propagate the semantic information correctly. Figure~\ref{fig:optflow_comp} shows the comparison between \gls{dis}~\cite{kroegerECCV2016} at a fast operating point and Gunnar-Farneback~\cite{DBLP:conf/scia/Farneback03} with a 2 layers pyramid, an averaging window of 9 pixels and 15 iterations. There is a substantial difference in the mIoU already after the first propagation. 

Experiments with higher quality settings for \gls{dis}~\cite{kroegerECCV2016} showed marginal improvements (below 0.2\%) on the mIoU even at high resolution, which motivated our choice for a faster operating point. With the ultra fast setting, the drop per propagation on the highest resolution is between 1.0\% and 1.5\% (1.2\% on average). This drop also tends to decrease when the resolution decreases, which is particularly interesting for the predictions and low level forward propagation of the features, since they operate at a resolution of $512\times256$ and $128\times64$ respectively.

\subsubsection{Uncertainties across the Evaluation Set}

\begin{figure}[t]
\begin{center}
\includegraphics[width=\linewidth]{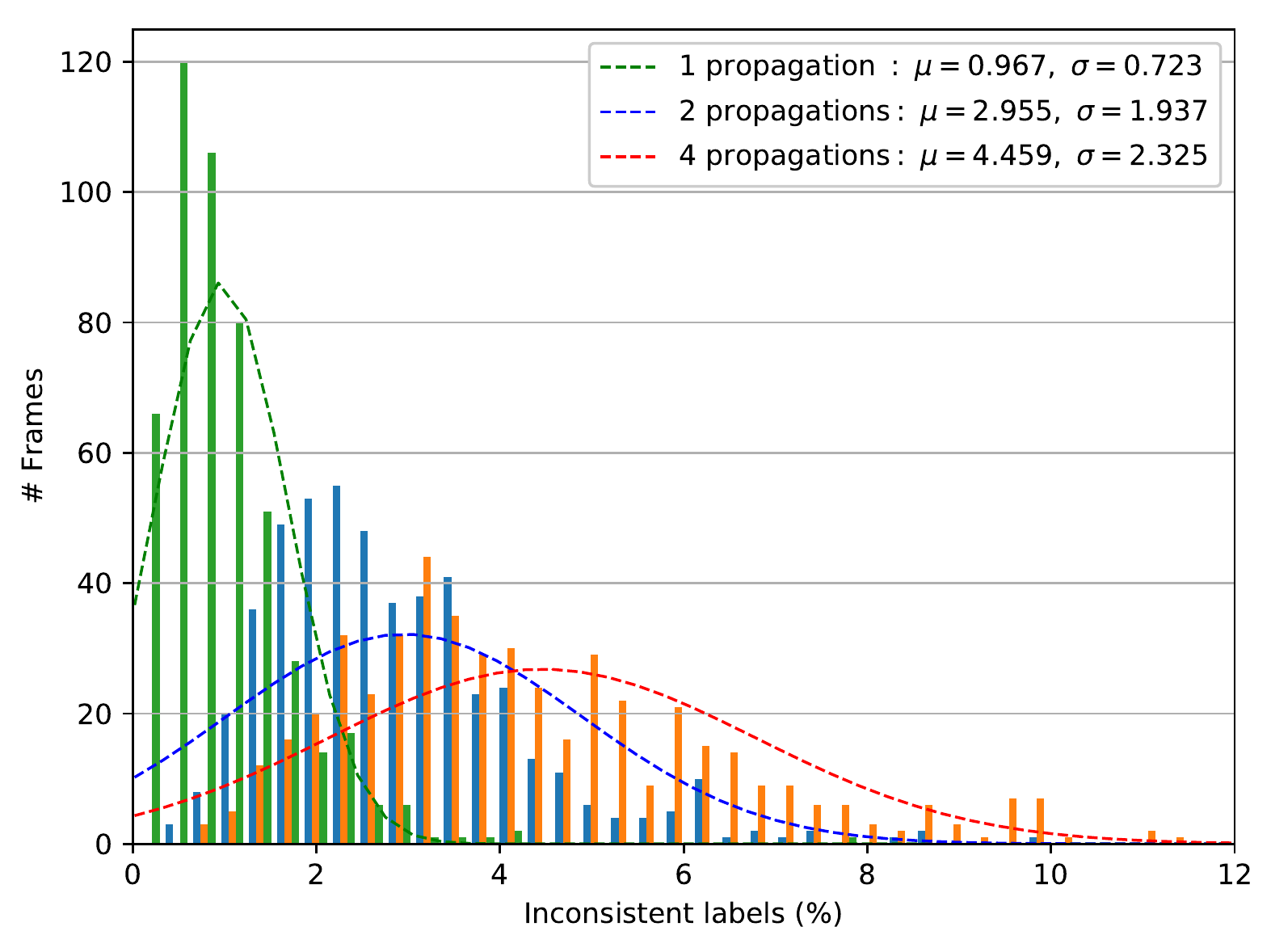}
\end{center}
   \caption{Re-partition of frames on the Cityscapes~\cite{Cordts2016Cityscapes} evaluation set as a function of their percentage of forward-backward inconsistent pixels from the optical flow, after 1, 2 and 4 propagations.}
\label{fig:optflow_bckfrd_uncert}
\end{figure}

The inconsistencies detected by forward-backward warping of the labels with the optical flow vary depending on the frame content and increase globally after each propagation. Figure~\ref{fig:optflow_bckfrd_uncert} shows on the evaluation set of CityScapes~\cite{Cordts2016Cityscapes} how the uncertainties are distributed depending on the number of propagations. This shows that even after 4 propagations, less than 5\% of the flow computed is detected as inconsistent on average. For frame-to-frame propagation, this drops to less than 1\%, which confirms that the optical flow is highly consistent.

\definecolor{Gray}{gray}{0.85}
\newcolumntype{a}{>{\columncolor{Gray}}c}
\definecolor{LightCyan}{rgb}{0.88,1,1}
\newcolumntype{b}{>{\columncolor{LightCyan}}c}

\begin{table*}[t]
	\footnotesize
	%\scriptsize
	%\tiny
	\setlength{\tabcolsep}{3.1pt}
	\begin{center}
		\begin{tabular}{| l | l@{\extracolsep{\fill}} | c c c c c a a a c c c a a c c c c c a |}
			\hline
			%\toprule[1pt]
			Method  & Total & road & swalk & build. & wall & fence & pole & tlight & sign & veg. & terrain & sky & person & rider & car & truck & bus & train & mbike & bike \\
			\hline\hline
			Baseline & 67.3 & 97.4 & 79.5 & 89.4 & 49.1 & 51.7 & 46.1 & 47.9 & 61.1 & 90.3 & 58.6 & 93.4 & 69.9 & 43.3 & 91.4 & 64.8 & 75.8 & 59.9 & 43.9 & 65.4 \\
			\hline\hline
			EVS 03 & 66.2 & 97.3 & 78.9 & 89.1 & 51.1 & 52.1 & 41.5 & 46.6 & 60.4 & 89.8 & 59.2 & 93.1 & 66.9 & 41.7 & 90.5 & 64.1 & 75.2 & 52.1 & 44.5 & 64.2\\
			%Warp loss & \textbf{-1.1$\color{red}\bigtriangledown$} &  -0.1 & -0.6 & -0.3 & +2.0 & +0.4 & \textbf{-4.6} & \textbf{-1.3} & \textbf{-0.7} & -0.5 & +0.6 & -0.3 & \textbf{-3.0} & \textbf{-1.6} & -0.9 & -0.7 & -0.6 & -7.8 & +0.6 & \textbf{-1.2}  \\
			EVS 02 & 66.8 & 97.3 & 79.1 & 89.3 & 51.0 & 52.3 & 44.5 & 47.1 & 61.4 & 90.2 & 59.5 & 93.2 & 69.0 & 42.5 & 90.9 & 63.9 & 75.5 & 52.5 & 45.0 & 64.9\\
			 Recovery & \textbf{+0.6$\color{green}\bigtriangleup$} & = & +0.2 & +0.2 & -0.1 & +0.2 & \textbf{+3.0} & \textbf{+0.5} & \textbf{+1.0} & +0.4 & +0.3 & +0.1 & \textbf{+2.1} & \textbf{+0.8} & +0.4 & -0.2 & +0.3 & +0.4 & +0.5 & \textbf{+0.7}  \\
			\hline\hline
			EVS 07 & 62.2 & 96.8 & 77.1 & 87.4 & 50.6 & 49.5 & 31.3 & 43.5 & 55.5 & 87.9 & 55.9 & 92.6 & 57.9 & 35.4 & 87.2 & 59.8 & 72.0 & 41.8 & 40.3 & 58.5 \\
			%Warp loss & \textbf{-5.1$\color{red}\bigtriangledown$} &  -0.6 & -2.4 & -2.0 & +1.5 & -2.2 & \textbf{-14.8} & \textbf{-4.4} & \textbf{-5.6} & -2.4 & -2.7 & -0.8 & \textbf{-12.0} & \textbf{-7.9} & -4.2 & -5.0 & -3.8 & -18.1 & -3.6 & \textbf{-6.9}\\
			EVS 06 & 63.0 & 96.6 & 75.6 & 87.9 & 50.2 & 49.8 & 36.0 & 44.4 & 57.5 & 89.1 & 56.6 & 93.2 & 63.6 & 36.5 & 87.5 & 59.2 & 70.8 & 41.5 & 41.1 & 60.0\\
			Recovery & \textbf{+0.8$\color{green}\bigtriangleup$} & -0.2 & -1.5 & +0.5 & -0.4 & +0.3 & \textbf{+4.7} & \textbf{+0.9} & \textbf{+2.0} & +1.2 & +0.7 & +0.6 & \textbf{+5.7} & \textbf{+1.1} & +0.3 & -0.6 & -1.2 & -0.3 & +0.8 & \textbf{+1.5}\\
            \hline
			%\bottomrule[1pt]
		\end{tabular}
	\end{center}
	\caption{Per-class results on Cityscapes after propagating 1 or 4 times the labels with refinement (EVS 02 and EVS 06) or without (EVS 03 and EVS 07). The corresponding recovery achieved by the Refiner is explicitely mentioned for both cases.}
	\label{tab:class_refiner_vs_flow}
\end{table*}

\subsection{\acrshort{evs} Operating Points}

\subsubsection{Per Class Impact of Warping and Refinement}

As discussed before, a simple forward mapping of the predictions made by the segmentation network can bring an important speedup factor, at a cost of an overall marginally degraded segmentation quality. Although the drop in mIoU per propagation might seem marginal, it is directly correlated to the mistakes of the optical flow (especially around boundaries of moving objects, thin objects and occlusions that may happen over time) and might have a big impact locally.

A per class analysis (see Table~\ref{tab:class_refiner_vs_flow}) confirms that the errors due to optical flow propagation affect most classes only marginally (below 1\% drop in IoU). Some of them are particularly affected by the wrong labeling: static thin objects (poles, traffic signs, traffic lights) and humans/small moving objects (person, rider, bike) are the most impacted classes (between 1\% and 5\% drop in IoU). 

\begin{figure*}[t]
\begin{center}

\begin{subfigure}{0.5\columnwidth}
\includegraphics[width=\linewidth]{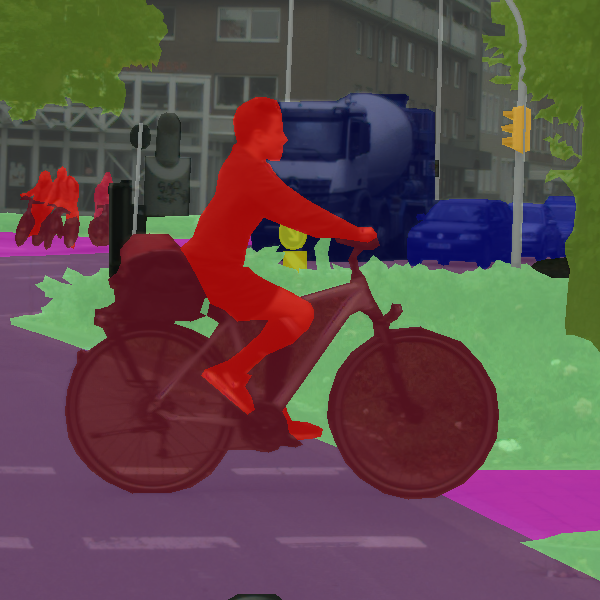}
\end{subfigure}
\begin{subfigure}{0.5\columnwidth}
\includegraphics[width=\linewidth]{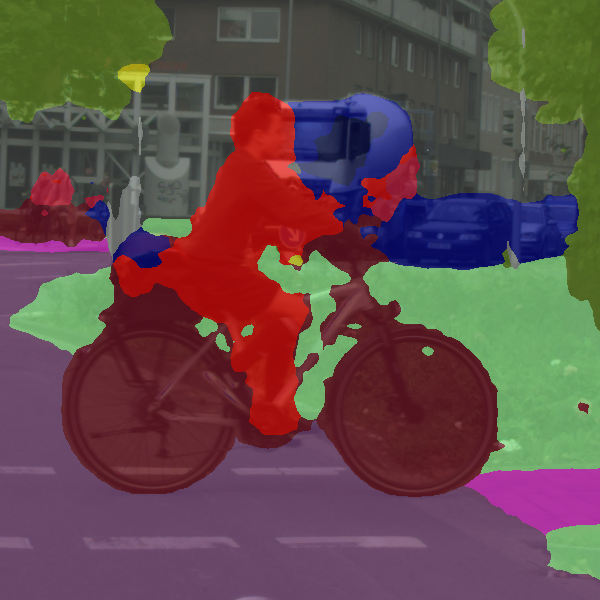}
\end{subfigure}
\begin{subfigure}{0.5\columnwidth}
\includegraphics[width=\linewidth]{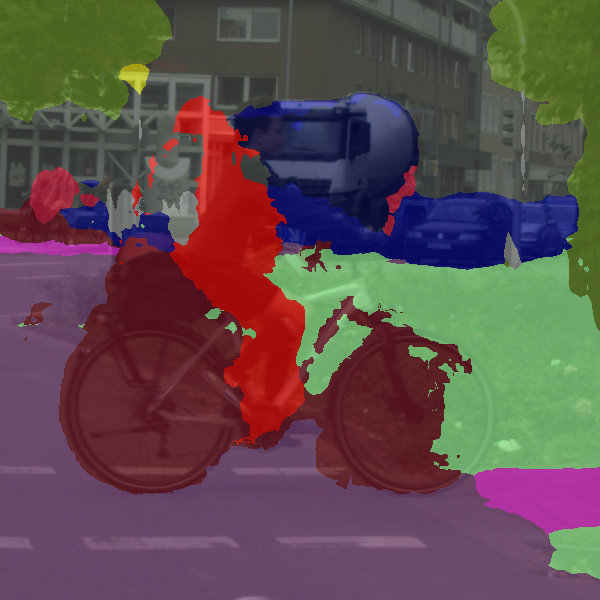}
\end{subfigure}
\begin{subfigure}{0.5\columnwidth}
\includegraphics[width=\linewidth]{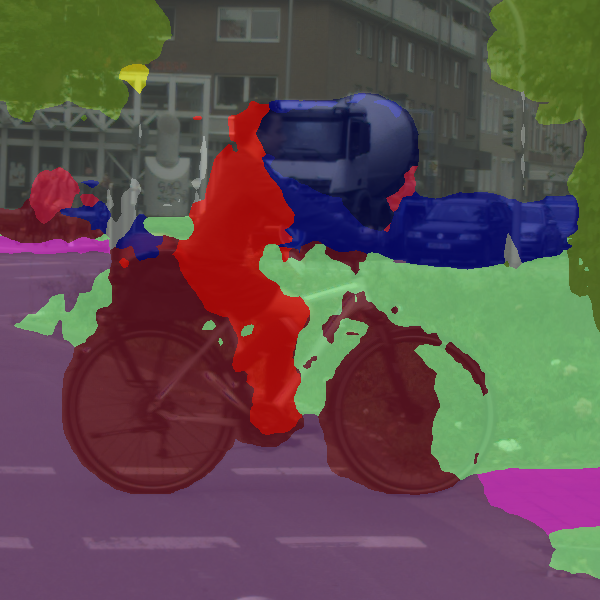}
\end{subfigure}\\[1em]
\begin{subfigure}{0.5\columnwidth}
\includegraphics[width=\linewidth]{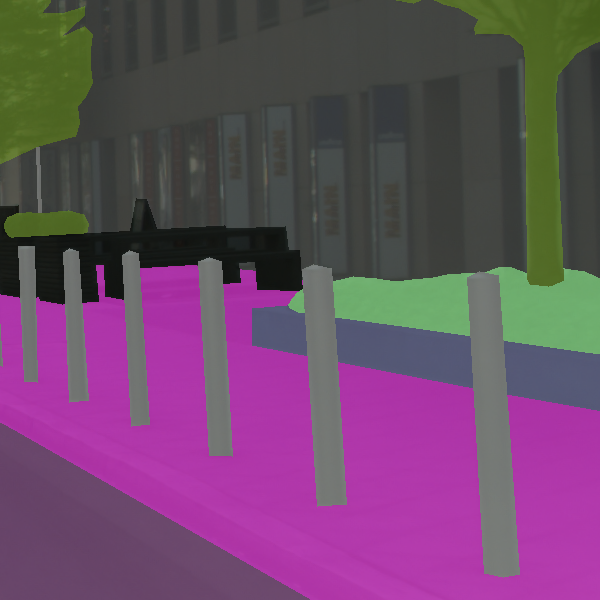}
\end{subfigure}
\begin{subfigure}{0.5\columnwidth}
\includegraphics[width=\linewidth]{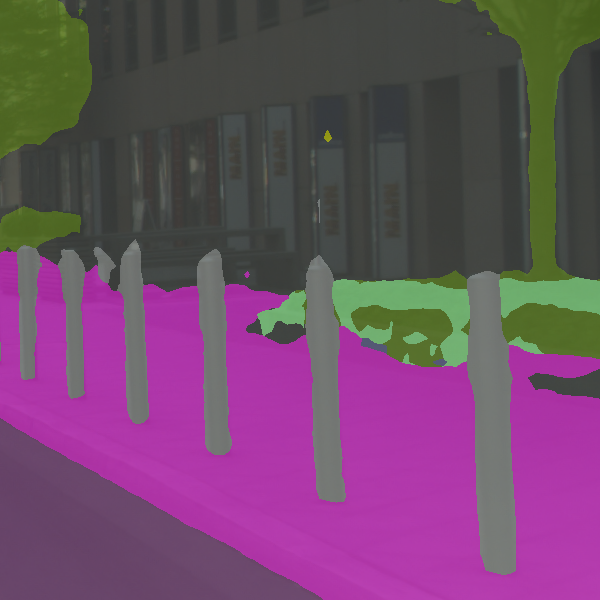}
\end{subfigure}
\begin{subfigure}{0.5\columnwidth}
\includegraphics[width=\linewidth]{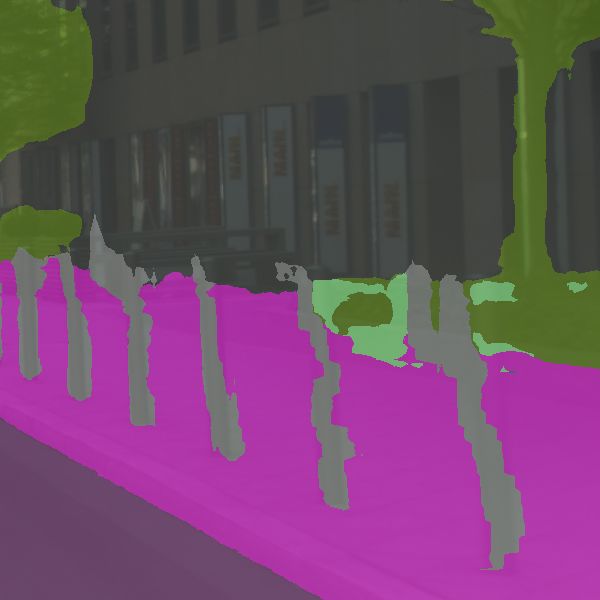}
\end{subfigure}
\begin{subfigure}{0.5\columnwidth}
\includegraphics[width=\linewidth]{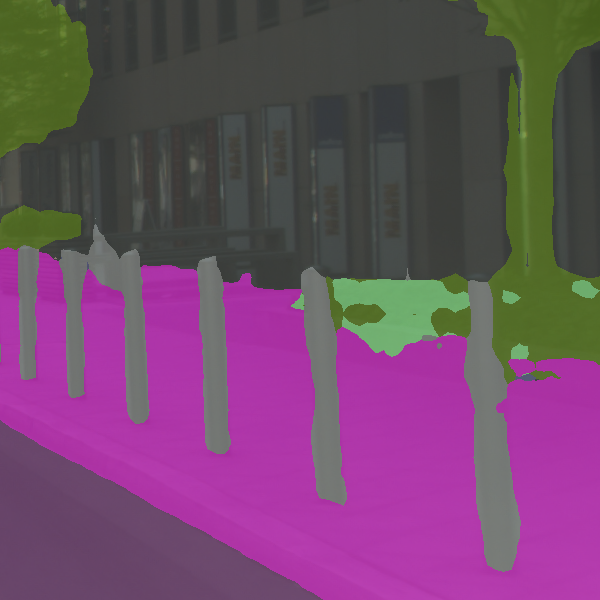}
\end{subfigure}\\[1em]
\begin{subfigure}{0.5\columnwidth}
\includegraphics[width=\linewidth]{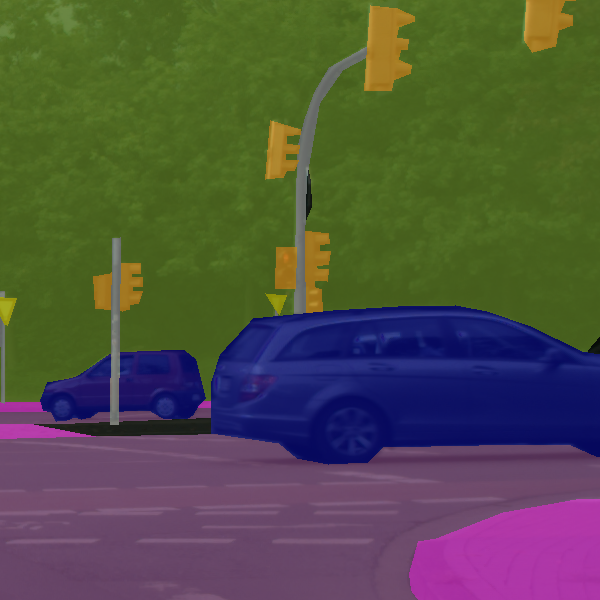}
    \caption{Ground truth}\label{fig:vis_groundtruth}
\end{subfigure}
\begin{subfigure}{0.5\columnwidth}
\includegraphics[width=\linewidth]{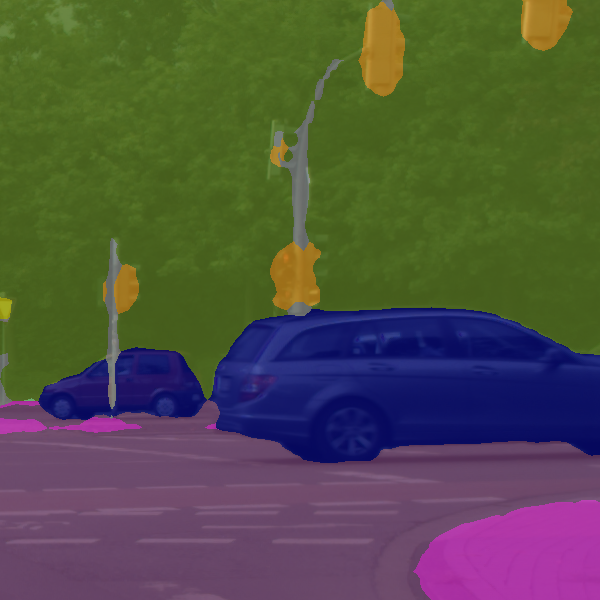}
    \caption{ICNet}\label{fig:vis_baseline}
\end{subfigure}
\begin{subfigure}{0.5\columnwidth}
\includegraphics[width=\linewidth]{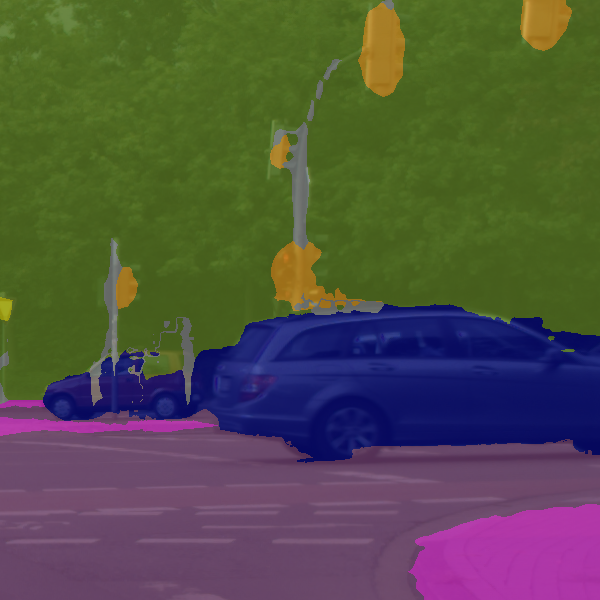}
    \caption{\gls{evs} Propagated}\label{fig:vis_propagated}
\end{subfigure}
\begin{subfigure}{0.5\columnwidth}
\includegraphics[width=\linewidth]{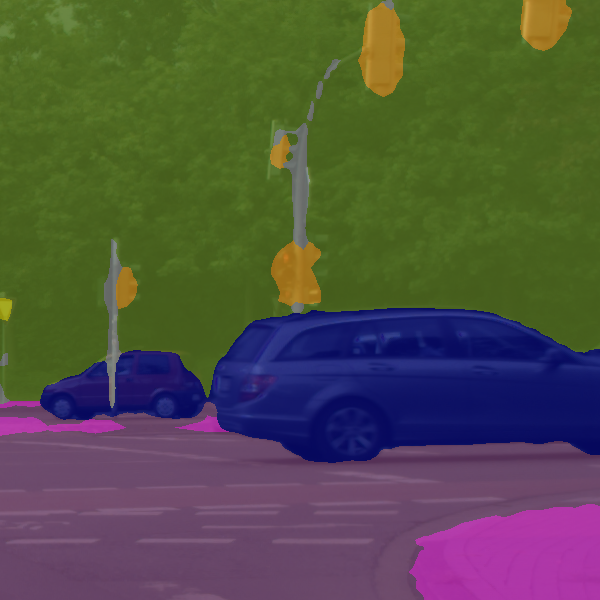}
    \caption{\gls{evs} Propagated + Refined}\label{fig:vis_refined}               
\end{subfigure}
\caption{Benefits of the Refiner on the propagated predictions in three problematic situations for the optical flow: a person riding a bike, thin poles on the side walk and occlusions from cars.}
\label{fig:visual_refined}
\end{center}
\end{figure*}

The Refiner is able to recover a large portion of the drop in IoU observed for those classes, especially after 1 frame propagation for poles, street signs and persons, even though the overall IoU gain is between 0.5\% and 1\%. Figure~\ref{fig:visual_refined} shows these typical situations where the refinement has a clear visible impact on these specific classes and shows that our Refiner can recover missing parts:
\begin{itemize}  
 \item[-] Thin objects such as poles, street signs or traffic lights are not always captured or heavily distorted by the camera motion.
 \item[-] Pedestrians on a crossing or cyclists and bikes are sometimes difficult to be fully captured with optical flow.
 \item[-] Missing parts due to occlusion and moving objects: a cyclist and bike passing in front of vegetation or two cars at a crossing.
\end{itemize}

Interestingly, the analysis also reveals that large static classes benefit from propagation (wall, fence, terrain) such that the IoU for these classes is higher than the IoU produced by the baseline, even without refinement (between 0.5\% and 2\% gain in IoU).

\subsubsection{Operating Point Comparison}

The structure of our \gls{evs} pipeline is defined by four parameters: the downscaling factor used by the segmentation network (\textbf{D}), the rate (every $n^{th}$ frame) at which the full segmentation is computed (\textbf{S}), warping (\textbf{W}) and refinement (\textbf{R}).
Table~\ref{tab:op_points} summarizes these operating points and compares them with state-of-the-art methods in terms of accuracy, frame rate and speedup factor compared to our baseline ICNet~\cite{Zhao_2018_ECCV}.

\begin{table}[t]
\begin{center}
\resizebox{1.\columnwidth}{!}{\begin{tabular}{|c|l|rlccccl|}
  \hline
   & Method & Framerate & Speedup & D & S & W & R & mIoU \\
  \hline\hline
  
  % Segm every 17 frame (16 prop), with middle branch res / 2  1000 / ((14.77+1.5) / 17)
  \multirow{21}{*}{\rotatebox[origin=c]{90}{Video pipeline}}
   &  EVS 14 (Ours) & 1045 Hz & $\times27.1$ & 0.5 & 17 & \cmark & \xmark & 46.0\% \\
  % Segm every 10 frame (0 prop), with middle branch res / 2  1000 / (14.77 / 10)
   & EVS 13 (Ours) & 677 Hz & $\times17.6$ & 0.5 & 10 & \xmark & \xmark & 35.3\% \\ 
  % Segm every 10 frame (9 prop), with middle branch res / 2  1000 / ((14.77+1) / 10)
   &  EVS 12 (Ours) & 634 Hz & $\times16.5$ & 0.5 & 10 & \cmark & \xmark & 50.7\% \\
  % Segm every 10 frame (0 prop). 1000 / (25.83 / 10)
   & EVS 11 (Ours) & 387 Hz & $\times10.1$ & 1.0 & 10 & \xmark & \xmark & 36.7\% \\
  % Segm every 10 frame (9 prop). 1000 / ((25.83+1) / 10)
   & EVS 10 (Ours) & 372 Hz & $\times9.7$ & 1.0 & 10 & \cmark & \xmark & 56.2\% \\
  % Segm every 5 frame (0 prop) with middle branch res / 2 1000 / ((14.77) / 5)
   & EVS 09 (Ours) & 339 Hz & $\times8.8$ & 0.5 & 5 & \xmark & \xmark & 42.8\% \\
  % Segm every 5 frame (0 prop). 1000 / ((25.83) / 5)
   & EVS 08 (Ours) & 192 Hz & $\times5.0$ & 1.0 & 5 & \xmark & \xmark & 46.4\% \\
  % Segm every 5 frame (4 prop). 1000 / ((25.83+0.5) / 5)
   & EVS 07 (Ours) & 190 Hz & $\times4.9$ & 1.0 & 5 & \cmark & \xmark & 62.2\% \\ 
  % Segm every 5 frame (4 prop) With refinement last. 1000 / ((25.83+0.5 + 1 + 13.6) / 5)
   & EVS 06 (Ours) & 122 Hz & $\times3.2$ & 1.0 & 5 & \cmark & \cmark & 63.0\% \\ 
  % Segm every 4 frame (3 prop) With refinement last. 1000 / ((25.83+0.3 + 1 + 13.6) / 4)
  %F04-WRW & 63.0\ \%$ & 10.2 ms & 98 Hz & $\times2.5$  \\
  % Segm every 3 frame (2 prop) 1000 / ((25.83 + 2*0.1) / 3)
   &  EVS 05 (Ours) & 115 Hz & $\times3.0$ & 1.0 & 3 & \cmark & \xmark & 64.7\% \\
  % Segm every 3 frame (2 prop) with refinement 1000 / ((25.83 + 2*0.1 + 1 + 13.6) / 3)
   & EVS 04 (Ours) & 74 Hz & $\times1.9$ & 1.0 & 3 & \cmark & \cmark & 65.6\% \\
  % Segm every 2 frame (1 prop) 1000 / ((25.83 + 2*0.1) / 2)
   &  EVS 03 (Ours) & 77 Hz & $\times2.0$ & 1.0 & 2 & \cmark & \xmark & 66.2\% \\
  % Segm every 2 frame (1 prop) with refinement 1000 / ((25.83 + 1*0.1 + 1 + 13.6) / 2)
   & EVS 02 (Ours) & 49 Hz & $\times1.2$ & 1.0 & 2 & \cmark & \cmark & 66.8\% \\
  % Segm every frame with ICNet + IAM. 1000 / (25.83 + 1)
   & EVS 01 (Ours) & 37 Hz & $\times0.95$ & 1.0 & 1 & \cmark & \cmark & 67.6\% \\
  %
  % 70.4% mIoU at 19.8 fps on the Cityscape dataset. HighSpeed: 30.4with 63.2% mIoU
  % Decision network that is trained and decides when to trigger keyframes
   & DVSN~\cite{Xu_2018_CVPR} & 30 Hz & $\times0.8$ & - & - & - & - & 62.6\%  \\
  % Best accuracy for alternating Schedule. 1000 / (60 + (18.7 + 23) / 2) ms 80.9 ms ~12Hz
  % => acceleration factor of 1.25 ?
   & Clockwork~\cite{10.1007/978-3-319-49409-8_69} & 12 Hz & $\times0.3$ & - & - & - & - & 64.4\%  \\
  % 360ms to151ms while decreas-ing the performance by4.9%. Using adaptive key frame
   & LLVS~\cite{Li_2018_CVPR} & 6.6 Hz & $\times0.2$ & - & - & - & - & 75.3\% \\
  % 66.9 - 69.2 accuracy    5.60fps On a K40. Maybe multiply by 2 to be fair ? $\times0.14$
   & DFF~\cite{zhu17dff} & 5.6 Hz & $\times0.1$ & - & - & - & - & 69.2\%  \\
  % Additional 75 - 335 ms (35 GRU unit) additional network for marginal improvements and temporal consist.
  % < 1% improvement with every backbone they tested.
   & GRFP~\cite{Nilsson_2018_CVPR} & 0.6 Hz & $\times0.02$ & - & - & - & - & 81.3\%  \\
  % +1.2 mIoU for 24 ms; also using DIS
   & NetWarp~\cite{gadde2017semantic} & 0.3 Hz & $\times0.01$ & - & - & - & - & 80.6\%  \\
  \hline\hline
  \multirow{6}{*}{\rotatebox[origin=c]{90}{Single frame}} & ENet~\cite{DBLP:journals/corr/PaszkeCKC16} & 77 Hz & $\times1.9$ & - & - & - & - & 58.3\% \\
  % ...
   & ERFNet~\cite{Romera2018ERFNetER} & 42 Hz & $\times1.1$ & - & - & - & - & 69.7\%  \\
  % Segm every frame with ICNet. 1000 / (25.83)
   & ICNet\cite{Zhao_2018_ECCV} & 39 Hz & Ref.-- & - & - & - & - & 67.3\% \\

   & SQ~\cite{Treml2016SpeedingUS} & 17 Hz & $\times0.4$ & - & - & - & - & 59.8\% \\
   & SegNet~\cite{Badrinarayanan2016SegNetAD} & 15 Hz & $\times0.4$ & - & - & - & - & 57.0\% \\
   & PSPNet~\cite{Zhao_2017_CVPR} & 0.8 Hz & $\times0.02$ & - & - & - & - & 81.2\% \\

  \hline
\end{tabular}}

\end{center}
\caption{Comparison of different \gls{evs} pipeline operating points. Numbers are reported on a Nvidia Titan Xp GPU and Intel Core i7-5930K CPU @3.50GHz for our pipeline and the reproduced ICNet~\cite{Zhao_2018_ECCV}. Numbers for other methods are reported from their respective papers on various hardware.}
\label{tab:op_points}
\end{table}

%------------------------------------------------------------------------
\section{Conclusion}

In this work, we introduce an \acrlong{evs} pipeline that pushes the boundaries of real-time video semantic segmentation in terms of computational efficiency by combining the benefits of deep \glspl{cnn} running on the GPU and a very fast optical flow running in parallel with the CPU. We propose different operating modes in order to focus either on frame rate or accuracy, from 67\% mIoU at $\sim40$  Hz to 46\% mIoU at $\sim1000$ Hz for $2048\times1024$ input images.

To compensate for the introduced errors in the predictions by the optical flow propagation around thin and/or moving objects (poles, persons or bikes), we propose a Refiner network to correct some errors and to generate a visually more appealing segmentation. The Refiner works with a dedicated \acrlong{iam} which focuses the prediction refinement on the relevant regions of the image. 

One of the strengths of our pipeline is that the segmentation network can be used as a black box method and can be replaced with any other segmentation network, bringing potentially more accuracy for the same speedups in the future. Moreover, our pipeline could benefit from a higher input frame rate because two consecutive frames are more similar and lead to a more accurate and reliable optical flow prediction (Cityscapes has a rather small frame rate of 17 Hz). Furthermore, the \gls{iam} introduced in this paper could be used in a future work as a way to dynamically adapt the behaviour of our pipeline depending to the input frames. In simple situations, the segmentation network and the Refiner could run less often such that the whole pipeline relies more on the optical flow when it is reliable. In more complex situations, the pipeline would then be able to force the re-segmentation more often to preserve a reasonable accuracy at the price of a lower frame rate.

\newpage

{\small
\bibliographystyle{ieee}
\bibliography{egbib}

\begin{thebibliography}{10}\itemsep=-1pt

\bibitem{Badrinarayanan2016SegNetAD}
V.~Badrinarayanan, A.~Kendall, and R.~Cipolla.
\newblock Segnet: A deep convolutional encoder-decoder architecture for image
  segmentation.
\newblock {\em IEEE Transactions on Pattern Analysis and Machine Intelligence},
  39:2481--2495, 2016.

\bibitem{Brostow2009SemanticOC}
G.~J. Brostow, J.~Fauqueur, and R.~Cipolla.
\newblock Semantic object classes in video: A high-definition ground truth
  database.
\newblock {\em Pattern Recognition Letters}, 30:88--97, 2009.

\bibitem{Brostow2008SegmentationAR}
G.~J. Brostow, J.~Shotton, J.~Fauqueur, and R.~Cipolla.
\newblock Segmentation and recognition using structure from motion point
  clouds.
\newblock In {\em ECCV}, 2008.

\bibitem{Butler:ECCV:2012}
D.~J. Butler, J.~Wulff, G.~B. Stanley, and M.~J. Black.
\newblock A naturalistic open source movie for optical flow evaluation.
\newblock In {A. Fitzgibbon et al. (Eds.)}, editor, {\em European Conf. on
  Computer Vision (ECCV)}, Part IV, LNCS 7577, pages 611--625. Springer-Verlag,
  Oct. 2012.

\bibitem{DBLP:journals/pami/ChenPKMY18}
L.~Chen, G.~Papandreou, I.~Kokkinos, K.~Murphy, and A.~L. Yuille.
\newblock Deeplab: Semantic image segmentation with deep convolutional nets,
  atrous convolution, and fully connected crfs.
\newblock {\em {IEEE} Trans. Pattern Anal. Mach. Intell.}, 40(4):834--848,
  2018.

\bibitem{Chen_2018_ECCV}
L.-C. Chen, Y.~Zhu, G.~Papandreou, F.~Schroff, and H.~Adam.
\newblock Encoder-decoder with atrous separable convolution for semantic image
  segmentation.
\newblock In {\em The European Conference on Computer Vision (ECCV)}, September
  2018.

\bibitem{Cheng_ICCV_2017}
J.~Cheng, Y.-H. Tsai, S.~Wang, and M.-H. Yang.
\newblock Segflow: Joint learning for video object segmentation and optical
  flow.
\newblock In {\em IEEE International Conference on Computer Vision (ICCV)},
  2017.

\bibitem{Cordts2016Cityscapes}
M.~Cordts, M.~Omran, S.~Ramos, T.~Rehfeld, M.~Enzweiler, R.~Benenson,
  U.~Franke, S.~Roth, and B.~Schiele.
\newblock The cityscapes dataset for semantic urban scene understanding.
\newblock In {\em Proc. of the IEEE Conference on Computer Vision and Pattern
  Recognition (CVPR)}, 2016.

\bibitem{DFIB15}
A.~Dosovitskiy, P.~Fischer, E.~Ilg, P.~H{\"a}usser, C.~Haz{\i}rba{\c{s}},
  V.~Golkov, P.~v.d. Smagt, D.~Cremers, and T.~Brox.
\newblock Flownet: Learning optical flow with convolutional networks.
\newblock In {\em IEEE International Conference on Computer Vision (ICCV)},
  2015.

\bibitem{DBLP:conf/scia/Farneback03}
G.~Farneb{\"{a}}ck.
\newblock Two-frame motion estimation based on polynomial expansion.
\newblock In J.~Big{\"{u}}n and T.~Gustavsson, editors, {\em Image Analysis,
  13th Scandinavian Conference, {SCIA} 2003, Halmstad, Sweden, June 29 - July
  2, 2003, Proceedings}, volume 2749 of {\em Lecture Notes in Computer
  Science}, pages 363--370. Springer, 2003.

\bibitem{floros12}
G.~Floros and B.~Leibe.
\newblock Joint 2d-3d temporally consistent semantic segmentation of street
  scenes.
\newblock 06 2012.

\bibitem{gadde2017semantic}
R.~Gadde, V.~Jampani, and P.~V. Gehler.
\newblock Semantic video cnns through representation warping.
\newblock In {\em IEEE International Conference on Computer Vision (ICCV)}.
  IEEE, 2017.

\bibitem{Geiger2013IJRR}
A.~Geiger, P.~Lenz, C.~Stiller, and R.~Urtasun.
\newblock Vision meets robotics: The kitti dataset.
\newblock {\em International Journal of Robotics Research (IJRR)}, 2013.

\bibitem{IMSKDB17}
E.~Ilg, N.~Mayer, T.~Saikia, M.~Keuper, A.~Dosovitskiy, and T.~Brox.
\newblock Flownet 2.0: Evolution of optical flow estimation with deep networks.
\newblock In {\em IEEE Conference on Computer Vision and Pattern Recognition
  (CVPR)}, 2017.

\bibitem{DBLP:journals/corr/abs-1901-02446}
A.~Kirillov, R.~B. Girshick, K.~He, and P.~Doll{\'{a}}r.
\newblock Panoptic feature pyramid networks.
\newblock {\em CoRR}, abs/1901.02446, 2019.

\bibitem{Krizhevsky:2012:ICD:2999134.2999257}
A.~Krizhevsky, I.~Sutskever, and G.~E. Hinton.
\newblock Imagenet classification with deep convolutional neural networks.
\newblock In {\em Proceedings of the 25th International Conference on Neural
  Information Processing Systems - Volume 1}, NIPS'12, pages 1097--1105, USA,
  2012. Curran Associates Inc.

\bibitem{kroegerECCV2016}
T.~Kroeger, R.~Timofte, D.~Dai, and L.~V. Gool.
\newblock Fast optical flow using dense inverse search.
\newblock In {\em Proceedings of the European Conference on Computer Vision
  ({ECCV})}, 2016.

\bibitem{10.1007/978-3-319-10599-4_45}
A.~Kundu, Y.~Li, F.~Dellaert, F.~Li, and J.~M. Rehg.
\newblock Joint semantic segmentation and 3d reconstruction from monocular
  video.
\newblock In D.~Fleet, T.~Pajdla, B.~Schiele, and T.~Tuytelaars, editors, {\em
  Computer Vision -- ECCV 2014}, pages 703--718, Cham, 2014. Springer
  International Publishing.

\bibitem{Lezama2011}
J.~Lezama, K.~Alahari, J.~Sivic, and I.~Laptev.
\newblock Track to the future: Spatio-temporal video segmentation with
  long-range motion cues.
\newblock In {\em IEEE Conference on Computer Vision and Pattern Recognition},
  2011.

\bibitem{Li_2017_CVPR}
X.~Li, Z.~Liu, P.~Luo, C.~Change~Loy, and X.~Tang.
\newblock Not all pixels are equal: Difficulty-aware semantic segmentation via
  deep layer cascade.
\newblock In {\em The IEEE Conference on Computer Vision and Pattern
  Recognition (CVPR)}, July 2017.

\bibitem{Li_2018_CVPR}
Y.~Li, J.~Shi, and D.~Lin.
\newblock Low-latency video semantic segmentation.
\newblock In {\em The IEEE Conference on Computer Vision and Pattern
  Recognition (CVPR)}, June 2018.

\bibitem{Liu2015SemanticIS}
Z.~Liu, X.~Li, P.~Luo, C.~C. Loy, and X.~Tang.
\newblock Semantic image segmentation via deep parsing network.
\newblock {\em 2015 IEEE International Conference on Computer Vision (ICCV)},
  pages 1377--1385, 2015.

\bibitem{DBLP:conf/cvpr/LongSD15}
J.~Long, E.~Shelhamer, and T.~Darrell.
\newblock Fully convolutional networks for semantic segmentation.
\newblock In {\em {IEEE} Conference on Computer Vision and Pattern Recognition,
  {CVPR} 2015, Boston, MA, USA, June 7-12, 2015}, pages 3431--3440. {IEEE}
  Computer Society, 2015.

\bibitem{DBLP:conf/ijcai/LucasK85}
B.~D. Lucas and T.~Kanade.
\newblock Optical navigation by the method of differences.
\newblock In A.~K. Joshi, editor, {\em Proceedings of the 9th International
  Joint Conference on Artificial Intelligence. Los Angeles, CA, USA, August
  1985}, pages 981--984. Morgan Kaufmann, 1985.

\bibitem{Mahasseni2017BudgetAwareDS}
B.~Mahasseni, S.~Todorovic, and A.~Fern.
\newblock Budget-aware deep semantic video segmentation.
\newblock {\em 2017 IEEE Conference on Computer Vision and Pattern Recognition
  (CVPR)}, pages 2077--2086, 2017.

\bibitem{Mehta_2018_ECCV}
S.~Mehta, M.~Rastegari, A.~Caspi, L.~Shapiro, and H.~Hajishirzi.
\newblock Espnet: Efficient spatial pyramid of dilated convolutions for
  semantic segmentation.
\newblock In {\em The European Conference on Computer Vision (ECCV)}, September
  2018.

\bibitem{Nilsson_2018_CVPR}
D.~Nilsson and C.~Sminchisescu.
\newblock Semantic video segmentation by gated recurrent flow propagation.
\newblock In {\em The IEEE Conference on Computer Vision and Pattern
  Recognition (CVPR)}, June 2018.

\bibitem{Ochs:2014:SMO:2693343.2693376}
P.~Ochs, J.~Malik, and T.~Brox.
\newblock Segmentation of moving objects by long term video analysis.
\newblock {\em IEEE Trans. Pattern Anal. Mach. Intell.}, 36(6):1187--1200, June
  2014.

\bibitem{DBLP:journals/corr/PaszkeCKC16}
A.~Paszke, A.~Chaurasia, S.~Kim, and E.~Culurciello.
\newblock Enet: {A} deep neural network architecture for real-time semantic
  segmentation.
\newblock {\em CoRR}, abs/1606.02147, 2016.

\bibitem{Perazzi2016}
F.~Perazzi, J.~Pont-Tuset, B.~McWilliams, L.~{Van Gool}, M.~Gross, and
  A.~Sorkine-Hornung.
\newblock A benchmark dataset and evaluation methodology for video object
  segmentation.
\newblock In {\em Computer Vision and Pattern Recognition}, 2016.

\bibitem{Richter_2016_ECCV}
S.~R. Richter, V.~Vineet, S.~Roth, and V.~Koltun.
\newblock Playing for data: {G}round truth from computer games.
\newblock In B.~Leibe, J.~Matas, N.~Sebe, and M.~Welling, editors, {\em
  European Conference on Computer Vision (ECCV)}, volume 9906 of {\em LNCS},
  pages 102--118. Springer International Publishing, 2016.

\bibitem{Romera2018ERFNetER}
E.~Romera, J.~M. Alvarez, L.~M. Bergasa, and R.~Arroyo.
\newblock Erfnet: Efficient residual factorized convnet for real-time semantic
  segmentation.
\newblock {\em IEEE Transactions on Intelligent Transportation Systems},
  19:263--272, 2018.

\bibitem{Sengupta2013Urban3S}
S.~Sengupta, E.~Greveson, A.~Shahrokni, and P.~H.~S. Torr.
\newblock Urban 3d semantic modelling using stereo vision.
\newblock {\em 2013 IEEE International Conference on Robotics and Automation},
  pages 580--585, 2013.

\bibitem{10.1007/978-3-319-49409-8_69}
E.~Shelhamer, K.~Rakelly, J.~Hoffman, and T.~Darrell.
\newblock Clockwork convnets for video semantic segmentation.
\newblock In G.~Hua and H.~J{\'e}gou, editors, {\em Computer Vision -- ECCV
  2016 Workshops}, pages 852--868, Cham, 2016. Springer International
  Publishing.

\bibitem{Shi:2000:NCI:351581.351611}
J.~Shi and J.~Malik.
\newblock Normalized cuts and image segmentation.
\newblock {\em IEEE Trans. Pattern Anal. Mach. Intell.}, 22(8):888--905, Aug.
  2000.

\bibitem{Tokmakov17}
P.~Tokmakov, K.~Alahari, and C.~Schmid.
\newblock Learning motion patterns in videos.
\newblock In {\em CVPR}, 2017.

\bibitem{tokmakov:hal-01511145}
P.~Tokmakov, K.~Alahari, and C.~Schmid.
\newblock {Learning Video Object Segmentation with Visual Memory}.
\newblock In {\em {ICCV - IEEE International Conference on Computer Vision}},
  pages 4491--4500, Venice, Italy, Oct. 2017. {IEEE}.

\bibitem{Treml2016SpeedingUS}
M.~Treml, J.~A. Arjona-Medina, T.~Unterthiner, R.~Durgesh, F.~Friedmann,
  P.~Schuberth, A.~Mayr, M.~Heusel, M.~Hofmarcher, M.~Widrich, B.~Nessler, and
  S.~Hochreiter.
\newblock Speeding up semantic segmentation for autonomous driving.
\newblock In {\em NIPS workshop}, 2016.

\bibitem{DBLP:journals/corr/WuSH16e}
Z.~Wu, C.~Shen, and A.~van~den Hengel.
\newblock Wider or deeper: Revisiting the resnet model for visual recognition.
\newblock {\em CoRR}, abs/1611.10080, 2016.

\bibitem{zhu17dff}
J.~D. L. Y. Y.~W. Xizhou~Zhu, Yuwen~Xiong.
\newblock Deep feature flow for video recognition.
\newblock 2017.

\bibitem{Xu_2018_CVPR}
Y.-S. Xu, T.-J. Fu, H.-K. Yang, and C.-Y. Lee.
\newblock Dynamic video segmentation network.
\newblock In {\em The IEEE Conference on Computer Vision and Pattern
  Recognition (CVPR)}, June 2018.

\bibitem{Zhao_2018_ECCV}
H.~Zhao, X.~Qi, X.~Shen, J.~Shi, and J.~Jia.
\newblock Icnet for real-time semantic segmentation on high-resolution images.
\newblock In {\em The European Conference on Computer Vision (ECCV)}, September
  2018.

\bibitem{Zhao_2017_CVPR}
H.~Zhao, J.~Shi, X.~Qi, X.~Wang, and J.~Jia.
\newblock Pyramid scene parsing network.
\newblock In {\em The IEEE Conference on Computer Vision and Pattern
  Recognition (CVPR)}, July 2017.

\end{thebibliography}
}

\end{document}